\documentclass[sigconf]{acmart}
\AtBeginDocument{%
  \providecommand\BibTeX{{%
    Bib\TeX}}}
\usepackage[capitalize,noabbrev]{cleveref}

\definecolor{uptoken}{HTML}{A8DA91}
\definecolor{downtoken}{HTML}{7FA6D3}
\definecolor{fluctoken}{HTML}{FAE196}
\definecolor{retainall}{HTML}{9778BE}
\usepackage[table,xcdraw]{xcolor}
\usepackage{multirow}
\copyrightyear{2026}
\acmYear{2026}
\setcopyright{cc}
\setcctype{by}
\acmConference[MM '26] {Proceedings of the 34th ACM International Conference on Multimedia}{November 10--14, 2026}{Rio de Janeiro, Brazil.}
\acmBooktitle{Proceedings of the 34th ACM International Conference on Multimedia (MM '26), November 10--14, 2026, Rio de Janeiro, Brazil}
\acmISBN{979-8-4007-2213-4/2026/11}
\acmDOI{10.1145/3767308.3834978}

\begin{document}

\title{Capturing Token Tendencies for Training-Free Token Pruning in Multimodal Large Language Models}

\author{Jie Ma}
\email{jiema100@stu.xmu.edu.cn}
\affiliation{%
  \institution{Key Laboratory of Multimedia Trusted Perception and Efficient Computing, Ministry of Education of China, Xiamen University}
  \city{Xiamen}
  \state{Fujian}
  \country{China}
}

\author{Zhike Qiu}
\email{zhikeqiu@outlook.com}
\affiliation{%
  \institution{Key Laboratory of Multimedia Trusted Perception and Efficient Computing, Ministry of Education of China, Xiamen University}
  \city{Xiamen}
  \state{Fujian}
  \country{China}
}

\author{Jie Gao}
\email{jiegao@stu.xmu.edu.cn}
\affiliation{%
  \institution{Key Laboratory of Multimedia Trusted Perception and Efficient Computing, Ministry of Education of China, Xiamen University}
  \city{Xiamen}
  \state{Fujian}
  \country{China}
}

\author{Jiayi Ji}
\email{jjyxmu@gmail.com}
\affiliation{%
  \institution{Key Laboratory of Multimedia Trusted Perception and Efficient Computing, Ministry of Education of China, Xiamen University}
  \city{Xiamen}
  \state{Fujian}
  \country{China}
}

\author{Qian Chen}
\email{chenqian@xmoc.edu.cn}
\affiliation{%
  \institution{School of Information Engineering, Xiamen Ocean Vocational College}
  \city{Xiamen}
  \state{Fujian}
  \country{China}
}

\author{Xiaoshuai Sun}
\authornote{Corresponding author.}
\email{xssun@xmu.edu.cn}
\affiliation{%
  \institution{Key Laboratory of Multimedia Trusted Perception and Efficient Computing, Ministry of Education of China, Xiamen University}
  \institution{Sino-Russian Research Center for Digital Economy}
  \city{Xiamen}
  \state{Fujian}
  \country{China}
}

\author{Rongrong Ji}
\email{rrji@xmu.edu.cn}
\affiliation{%
  \institution{Key Laboratory of Multimedia Trusted Perception and Efficient Computing, Ministry of Education of China, Xiamen University}
  \city{Xiamen}
  \state{Fujian}
  \country{China}
}
\renewcommand{\shortauthors}{Jie Ma et al.}

\begin{abstract}
     While visual token pruning is essential for efficient Multimodal Large Language Models (MLLMs), existing training-free methods suffer from a critical limitation: they rely on \textit{static, instantaneous heuristics} to perform \textit{irreversible} filtering. This approach ignores the hierarchical nature of MLLMs, where token importance often evolves dynamically rather than remaining fixed across layers. Consequently, tokens essential for deep-layer reasoning are often prematurely discarded by shallow-layer estimates. To address this, we propose Trend-aware Pruning, a novel framework that elevates pruning from a local snapshot decision to a temporal trajectory modeling problem. Instead of relying on isolated scores, our method captures the momentum of attention flow. This enables a dynamic rectification mechanism that selectively reactivates ``late-blooming'' tokens---those initially undervalued but exhibiting rising semantic importance---thereby preventing the loss of critical visual cues. Extensive experiments demonstrate that our approach achieves a superior efficiency-performance trade-off across diverse multimodal tasks. Notably, it reduces visual tokens by over 77.8\%, retaining only approximately 23 tokens in the final layer while maintaining competitive performance, offering a robust and reversible solution for high-efficiency multimodal inference. The project page is available at https://github.com/JieMaMagic/Trend-aware-Pruning
\end{abstract}




\maketitle

\section{Introduction}

\begin{figure}[!t]
\begin{center}
\centerline{\includegraphics[width=\columnwidth]{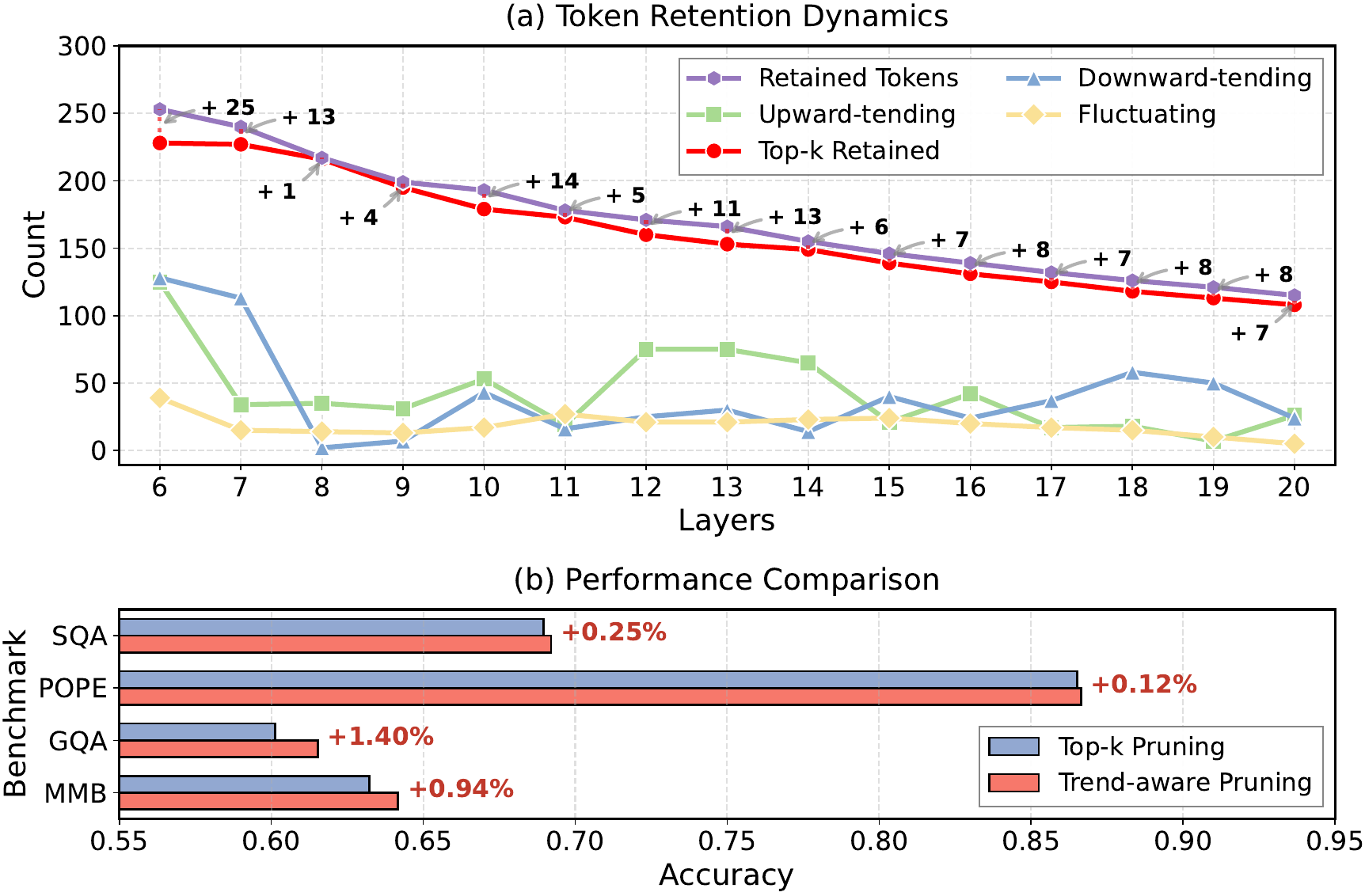}}

\caption{Token retention dynamics and performance comparison.
(a) Token Retention Dynamics (e.g., Layer 6$\sim$20). 
Our \textit{Trend-aware Pruning} (purple, “+” denotes recovered tokens) preserves informative visual tokens by modeling their tendencies, which are often discarded by static Top-$k$ pruning (red).
(b) Performance Comparison. Results on multimodal benchmarks show that our method consistently outperforms Top-$k$ pruning, with particularly strong gains on visual reasoning tasks such as GQA. This highlights the importance of modeling token tendencies and shows that single-layer importance is insufficient.}
\label{intro-trend-compare}
\end{center}
\end{figure}

Multimodal Large Language Models (MLLMs) have demonstrated impressive performance in a wide range of real-world applications, including robotics~\cite{Evo-1,EvoAgent_mm}, autonomous driving~\cite{AutonomousDriving,AD_mm}, and agent assistant systems~\cite{cogagent,multi_agent_mm}. By adopting vision encoders with large language models, MLLMs process complex multimodal inputs to achieve remarkable progress in vision-language understanding and reasoning. However, these advances  \cite{qwenvl25,llava,internvl2,deepseekai2025deepseekv3technicalreport} generate a large number of visual tokens when processing high-resolution images and multi-image inputs. Since the Attention mechanism has quadratic complexity, the large number of visual tokens leads to a high computational cost. This severe visual token redundancy has become a major bottleneck for inference.

To mitigate this computational burden, recent studies~\cite{Fastv,sparsevlm} have explored visual token pruning strategies to reduce the sequence length processed by the decoder. Most existing methods rely on static heuristics (e.g., Attention Top-$k$) to estimate token importance at an early stage, retaining only a fixed subset of tokens layer-by-layer. While effective in reducing FLOPs, these approaches treat pruning as a \textit{single-shot, irreversible operation}. Once a token is discarded based on a shallow-layer estimate, its information is permanently lost. This limitation raises a fundamental challenge: \textit{Is an early-layer decision reliable enough to determine the final semantic value of a visual token? If not, can we dynamically recover informative tokens to minimize semantic degradation?}

Our approach draws inspiration from human cognition~\cite{nerualaction,cellattention}, where attention is not a fixed snapshot but a dynamic process that adapts as the understanding evolves. Similarly, recent analyses~\cite{multimodelProcess,VisionFunctionLayer,sparsevlm,AutoPrune} reveal that MLLM layers play distinct functional roles: visual tokens initially represent low-level features (e.g., texture) in shallow layers and gradually evolve into high-level semantic entities in deeper layers. Consequently, a token that appears redundant in an early layer may become critical for reasoning in a deeper layer. Static pruning methods~\cite{Fastv,PyramidDrop,fasterVLM,holov,sparsevlm}, which ignore this evolution, are prone to prematurely discarding essential visual cues. This phenomenon is visualized in \cref{intro-trend-compare}. As shown in \cref{intro-trend-compare}(a), while standard methods enforce a strict decrease in token count, our approach identifies and ``rescues'' tokens that exhibit rising importance trends, thereby preserving critical semantics and boosting performance as reported in \cref{intro-trend-compare}(b).

To fundamentally address the misalignment between static pruning and dynamic semantic evolution, we propose \textbf{Trend-aware Pruning} method. Diverging from conventional snapshot-based approaches~\cite{Fastv, PyramidDrop}, we redefine visual token pruning as a temporal trajectory modeling problem. Specifically, our framework incorporates a Layer-wise Token Collector to aggregate historical attention states, capturing the \textit{momentum} of token importance rather than isolated estimates. Based on these trajectories, we employ Adaptive Flow Identification to categorize tokens into distinct evolutionary patterns (i.e., \textit{Upward}, \textit{Fluctuating}, and \textit{Downward} tendencies). Crucially, this drives our Flow Activation mechanism, which selectively reactivates tokens that exhibit positive semantic trends but are undervalued by static Top-$k$ selection. This strategy shifts the paradigm from irreversible filtering to dynamic, reversible selection, ensuring that critical visual cues are preserved even under aggressive sparsity constraints.

We conduct comprehensive experimental evaluations across multiple mainstream MLLMs benchmarks with varying parameter sizes (e.g., 0.5B, 7B and 13B) and architectures (LLaVA series \cite{llava,llavaOneVision,llavenext} and QwenVL \cite{qwenvl25}).
The experimental results show that our method has competitive performance compared to the state-of-the-art training-free token pruning methods. For instance, at a 50\% pruning ratio, our approach retains 98.89\% of the average performance while significantly reducing the computational FLOPs to 55.10\%. Notably, our method reduces the number of visual tokens input to the final layer to approximately 23, which is the lowest among all compared methods, demonstrating superior efficiency while maintaining highly competitive performance.


To summarize, our main contributions are as follows:

 \begin{itemize}
    \item We reformulate visual token pruning from a static single-shot operation into a dynamic cross-layer process, explicitly modeling the semantic evolution of visual tokens to capture trend information overlooked by existing methods.
    \item We propose a Trend-aware Pruning method that jointly guides token discarding and selective reactivation, enabling pruned tokens to be dynamically recovered when their semantic importance emerges, thus mitigating degradation from the irreversible visual cue loss.
    \item Extensive experiments across diverse MLLMs demonstrate that Trend-aware Pruning achieves a highly competitive efficiency–performance trade-off, preserving critical visual cues while significantly reducing computation under aggressive pruning.
\end{itemize}

\begin{figure*}[!t]
\begin{center}
\centerline{\includegraphics[width=1.0\linewidth]{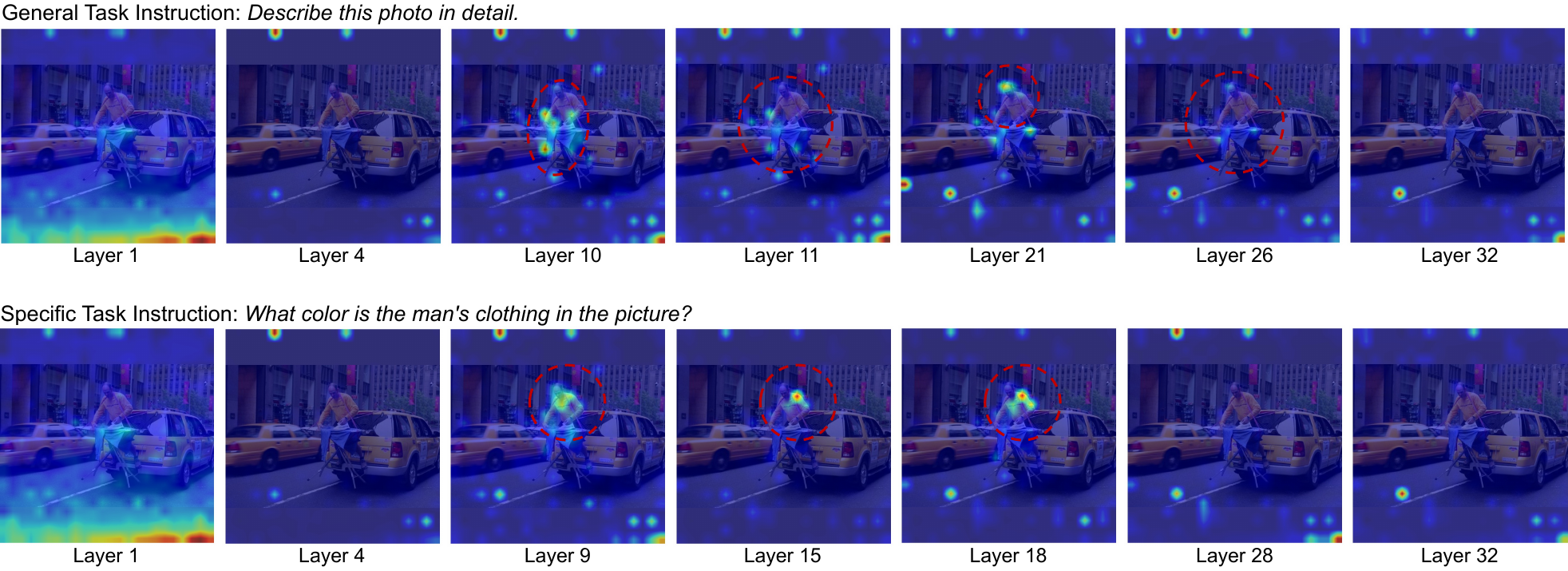}}
\caption{Visualization of layer-wise attention dynamics in LLaVA-v1.5-7B.  We compare the attention heatmaps of visual tokens across different layers for two distinct task types: a general description task (top) and a specific attribute inquiry task (bottom).
Red dashed circles highlight regions of visual attention, illustrating the heterogeneity in the model's focus across different layers. This layer-wise heterogeneity demonstrates that token importance is dynamic rather than static, indicating that pruning based on a single early layer risks discarding tokens that are critical for deep-layer reasoning. Best viewed in color.}
\label{fig:prem_heatmap}
\end{center}
\end{figure*}

\section{Related Work}

\subsection{Multimodal Large Language Models}

Multimodal Large Language Models aim to build unified architectures that integrate visual perception with language understanding and reasoning. A typical MLLM consists of a vision encoder \cite{clip,sigclip}, a projection module for modality alignment, and a large language model (LLM) \cite{llama} for multimodal reasoning.
Recent advanced MLLMs~\cite{llava,liu2024llavanext,qwenvl25,qwen3technicalreport,internvl2,deepseekai2025deepseekv3technicalreport} have significantly improved multimodal perception and reasoning capabilities. This progress has been driven by scaling up both model capacity and visual input resolution, leading to strong performance on multimodal benchmarks. In particular, high-resolution and tiling-based encoding strategies \cite{qwen3technicalreport,llavaOneVision} are widely adopted to decompose images into dense patches or multi-scale visual tokens, enabling the preservation of fine-grained visual details.
However, this paradigm inevitably produces extremely long visual token sequences, introducing substantial redundancy and imposing heavy computational and memory costs. These observations motivate the exploration of effective token reduction mechanisms that can identify informative visual cues, remove redundant tokens, and maintain performance.

\subsection{Visual Token Pruning}

Visual token pruning \cite{Fastv,fasterVLM,holov,sparsevlm,AutoPrune,PyramidDrop,ACCM,zoo_pruning,nuwa} has emerged as a mainstream technique to improve the efficiency of MLLMs, reducing visual redundancy and computational cost while maintaining multimodal performance. Early methods \cite{Fastv,fasterVLM,holov,sparsevlm} typically rely on attention scores and apply fixed pruning ratios to identify important tokens. While these strategies provide computational savings, they usually operate in a layer-isolated manner, assuming that token importance is locally decidable and temporally fixed. Such assumptions neglect the hierarchical and progressive formation of visual representations, limiting pruning effectiveness.
To overcome these limitations, recent adaptive pruning methods have been proposed \cite{AutoPrune,PyramidDrop,sparsevlm}. AutoPrune \cite{AutoPrune} provides a customized pruning policy that dynamically adapts to each input. PyramidDrop (PDrop) \cite{PyramidDrop} introduces a progressive reduction of tokens as the network depth increases, improving both training and inference efficiency. SparseVLM \cite{sparsevlm} leverages text-guided cues to adaptively determine the sparsification ratio. Although these approaches improve flexibility and efficiency, they lack an explicit modeling of cross-layer token dynamics and do not fully address the limitations of static or layer-isolated pruning.

\begin{figure*}[!ht]
\begin{center}
\centerline{\includegraphics[width=0.95\linewidth]{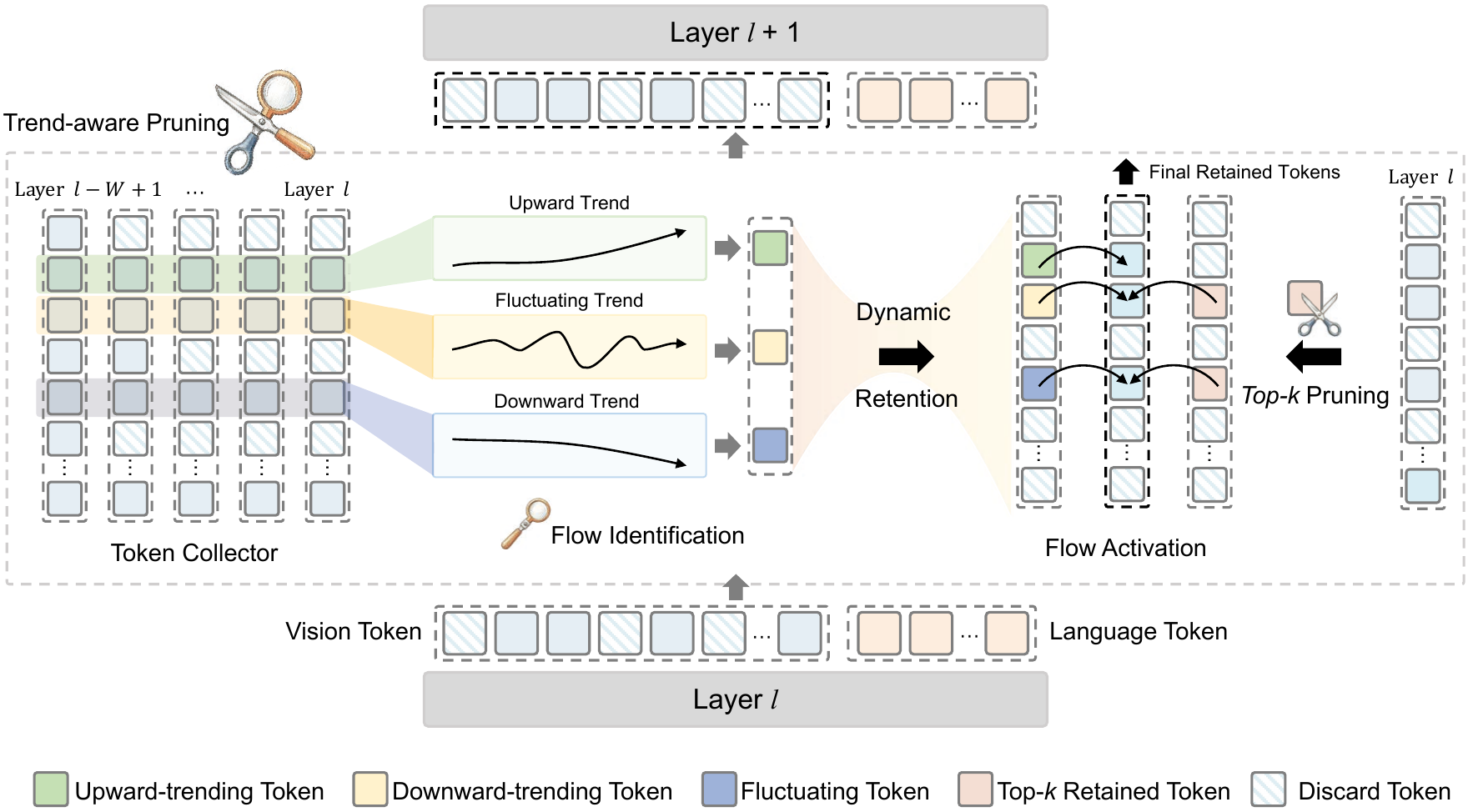}}
\caption{An overview of Trend-aware Pruning. The framework comprises three stages: (1) The Layer-wise Token Collector aggregates historical attention scores; (2) Flow Identification categorizes tokens into Upward (Green), Downward (Yellow), and Fluctuating (Blue) trends; (3) Flow Activation merges these dynamic tokens with static Top-$k$ (Pink) tokens to form the final preserved set for the next layer.}
\label{fig:framework}
\end{center}
\end{figure*}

\section{Methodology}

\label{sec-method}

In this section, we present Trend-aware Pruning, a dynamic framework for efficient MLLM inference. We commence by formulating the MLLM architecture and analyzing the layer-wise heterogeneity of attention patterns (\cref{sec-preliminaries}). Building on these insights, we detail the Layer-wise Token Collector and Attention Flow mechanism (\cref{subsec-collector}), followed by our strategies for adaptive flow identification (\cref{subsec-identification}) and dynamic token retention (\cref{subsec-activation}). We further provide a theoretical analysis of computational complexity (\cref{subsec-TheoreticalAnalysis}). The overall framework is shown in \cref{fig:framework}.

\subsection{Preliminaries}
\label{sec-preliminaries}
\textbf{Generic Architecture of MLLMs}. MLLMs typically follow a modular vision-language architecture composed of a pre-trained vision encoder $\mathcal{E}_v$, a multimodal alignment projector $\phi$, and a large language model $\mathcal{M}$. As a conditional probabilistic model, given a vision input $\mathbf{x}_v$ and a textual instruction $\mathbf{x}_t$ to generate a target response $\mathbf{y}$. Specifically, $\mathcal{E}_v$ extracts dense feature representations $\mathbf{H}_v$ from $\mathbf{x}_v$, which are subsequently mapped into the LLM's embedding space via $\phi$, yielding visual tokens $\mathbf{Z}_v = \phi(\mathcal{E}_v(\mathbf{x}_v))$. These visual tokens are concatenated with the tokenized text instruction prompt $\mathbf{Z}_t$ to form a unified input sequence of tokens $[\mathbf{Z}_v; \mathbf{Z}_t]$. The LLM architecture $\mathcal{M}$, serving as the core multimodal reasoning module, achieves implicit cross-modal fusion through attention mechanisms. Formed by $L$ stacked decoder layers, it employs an autoregressive process to maximize the likelihood $p(\mathbf{y} \mid \mathbf{x}_v, \mathbf{x}_t)$. 
Therefore, it is crucial to design a training-free, plug-and-play mechanism to effectively exploit the diverse hierarchical representations within MLLMs, thereby fully leveraging their intrinsic capabilities.

\begin{figure*}[!t]
\begin{center}
\centerline{\includegraphics[width=0.95\linewidth]{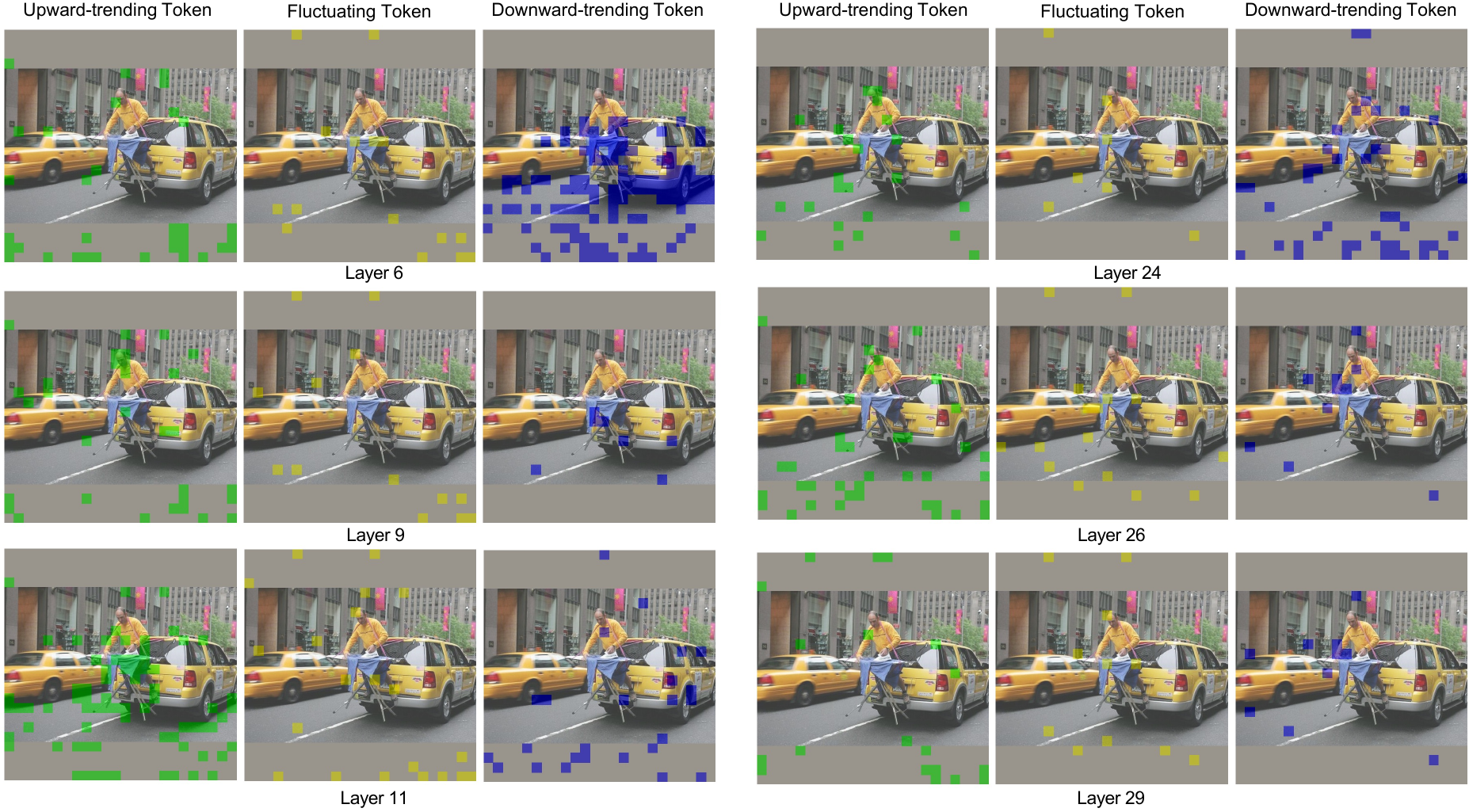}}
\caption{Visualization of trend-aware tokens via adaptive flow identification with instruction: \textit{`` What is unusual about this image? ''} in LLaVA-v1.5-7B.
We visualize tokens grouped into upward-trending, fluctuating, and downward-trending across layers. All three trends capture context-relevant tokens, while exhibiting distinct cross-layer retention patterns. This demonstrates that our trend-aware modeling characterizes diverse token contributions and enables dynamic token pruning.
}
\label{justification}
\end{center}
\end{figure*}

\noindent \textbf{Layer-wise Attention Heterogeneity.}
The internal processing dynamics in MLLMs \cite{multimodelProcess} identify a consistent stage-wise structure, where layers at different depths serve distinct functional roles, leading to pronounced layer-wise attention heterogeneity and representation shift.
We visualize the attention heatmaps of visual tokens across different layers in \cref{fig:prem_heatmap} under a general description task and a specific attribute query.
Early layers (e.g., Layer 1 and Layer 4) exhibit dispersed attention, with responses widely distributed across the road surface, vehicle, and background structures.  Pruning visual tokens based on such early responses would likely remove tokens that are weakly activated at shallow layers but become critical at deeper stages (e.g., human-centric regions that only dominate after Layer 15).
As depth increases (e.g., Layer 9–15), attention progressively filters out irrelevant background regions and begins to focus on salient objects, particularly the human and nearby vehicles. 
By deeper layers (e.g., Layer 21/26/28 and Layer 32), attention becomes concentrated, forming compact high response regions marked by the red dashed circles that align with task-relevant targets. Although both tasks consistently follow a \textit{dispersion-to-concentration trend}, their convergence patterns diverge. The general description task maintains a broader focus on multiple foreground objects, while the specific attribute query concentrates on the human body.

This observation highlights a key problem: \textit{relying on attention from just one layer for pruning is unreliable}. Each layer captures different types of information, from broad context in shallow layers to specific details in deeper ones. This means token importance is not fixed, which evolves with decoder layer depth, textual content, and task-specific demands. Therefore, these findings motivate cross-layer pruning mechanisms that dynamically integrate attention scores across layers to retain potentially important visual tokens.

\subsection{Token Collector \& Attention Flow}
\label{subsec-collector}

To capture this spatio-temporal dynamic and mitigate representation collapse caused by static pruning \cite{Fastv}, we introduce a Layer-wise Token Collector as a memory-based heuristic for modeling the visual token tendencies.

\noindent \textbf{Layer-wise Token Collector.} To overcome the static pruning methods that use the same sparsity rate for each layer and consider the vision representation of each layer to be the same. The layer-wise token collector maintains a short-term memory cache acting as a sliding observation window $\mathcal{W}_l$. Let $\mathbf{a}_l \in \mathbb{R}^{N^v_{l}}$ represent the aggregated attention score for $N^v_{l}$ visual tokens at the current layer $l$. The layer-wise token collector stores the sequence of historical attention states:
\begin{equation}
    \mathcal{W}_l = [\mathbf{a}_{l-W+1}, \dots, \mathbf{a}_{l-1}, \mathbf{a}_l],
\end{equation}
where $W$ is the window size. This token collector operates as a First-In-First-Out queue. To ensure cross layer token alignment, we further filter the historical attention cache using the currently selected token indices to guarantee consistent flow calculation for the next step.

\noindent \textbf{Modeling Token Flow as Trajectories.} To explicitly capture the tendency of visual tokens, we therefore compute the Token Flow $\mathcal{F}_l$, which serves as a quantitative metric for token dynamics by calculating the step-wise difference within the observation window:
\begin{align}
    \boldsymbol{\delta}_{k} &= \mathbf{a}_{k} - \mathbf{a}_{k-1}, \quad \forall k \in \{l-W+2, \dots, l\} \\
    \mathcal{F}_l &= \{ \boldsymbol{\delta}_{k} \}_{k=l-W+2}^{l},
\end{align}
where $k$ denotes the historical layer index. The discrete token trend $\boldsymbol{\delta}_{k}$ effectively decouples the representation change direction from its single-layer importance score. Crucially, this token flow serves as the foundational understanding for the Flow Activation mechanism. By identifying and “rescuing” tokens that currently important score low but exhibit considerable dynamic potential, it prevents prematurely discarding or fluctuating semantic vision cues.

\subsection{Adaptive Flow Identification}
\label{subsec-identification}

Considering the vision representation comprises a mixture of task-relevant semantic cues and stochastic noise. To determine which tokens to preserve, we firstly analyze the monotonicity of their attention flow. 
A stepwise increase in token flow indicates that a token is accumulating semantic importance for the current generation task. 
Conversely, a continuous decrease indicates a representation shift or irrelevance, and overlooking such trend tokens may potentially lead to the loss of critical visual cues.
Accordingly, we explicitly model both trends in token flow. Tokens with increasing attention are defined as Upward Trends $T_i^{\text{up}}$, while tokens with decreasing attention are defined as Downward Trends $T_i^{\text{down}}$. Formally, given token flow $\mathcal{F}_l$, the tendency consistency score is defined as:
\begin{align}
    T_i^{\text{up}} &= \frac{1}{W-1} \sum_{k=0}^{W-2} \mathbb{I}(\delta_{l-k, i} > 0) ,\\
    T_i^{\text{down}} &= \frac{1}{W-1} \sum_{k=0}^{W-2} \mathbb{I}(\delta_{l-k, i} < 0),
\end{align}
where $i \in \{1, 2, \dots, N^v_l\}$ represents the $i$-th token, $\mathbb{I}(\cdot)$ is the indicator function.  In addition to monotonic trends, some tokens exhibit fluctuating attention, reflecting visual cues whose relevance varies across layers. 
Discarding these tokens can lead to loss of informative visual cues. 
We then model fluctuating tokens separately, capturing their oscillating patterns and reduce relevant vision cues loss. We quantify this via token flow Volatility $T_i^{\text{fluct}}$:
\begin{equation}
   T_i^{\text{fluct}} = \sqrt{\frac{1}{W-2} \sum_{k=l-W+2}^{l} (\delta_{k, i} - \bar{\delta}_i)^2},
\end{equation}
where $\bar{\delta}_i$ is the mean of $ {\delta}_i$.

Subsequently, we employ a distribution-aware adaptive thresholding strategy to robustly retain the critical vision cues. Specifically, for given trend-aware scores $T\in \mathbb{R}^{N^v_l}$, the selection threshold $\tau$ is defined as:
\begin{equation}
   \tau (T, \lambda) = \mu_{T} + \lambda \cdot \sigma_{T},
\end{equation}
where $\mu_{T}$ and $\sigma_{T}$ represent the population statistics. $\lambda$ controls the sensitivity of trend variation and thus directly modulates pruning aggressiveness. Noted that, this adaptive strategy provide a dynamic threshold, rather than empirical settings.
It retains candidates $i$-th token where the $T_i$ satisfies  $(T_i - \mu_{T}) / \sigma_{T} > \lambda$.

\noindent \textbf{Justification.} To validate the effectiveness of the adaptive flow identification, we visualize trend-aware tokens by dynamically identifying upward-trending, fluctuating, and downward-trending. As shown in \cref{justification}, the flow identification effectively captures the most informative tokens. Tokens selected from early layers (e.g., Layer 9) to deeper layers (e.g., Layer 24 and 29) progressively focus on the instruction relevance regions to determine  ``\textit{unusual}'', such as human, nearby vehicles and road surface.
This layer-wise token selection adaptability ensures that the pruning mechanism retains contextually relevant tokens, enabling more efficient and dynamic token selection compared to static pruning strategies. Furthermore, this mitigates the “distraction” phenomenon where attention erroneously attends to irrelevant visual tokens.
By explicitly modeling token trend dynamics, our approach provides a novel pruning paradigm that captures both persistent and emerging informative tokens across the hierarchy, while effectively discarding redundant ones.

\begin{table*}[!t] 
    \centering
    \small
            \caption{Performance comparison with training-free token pruning methods on LLaVA-v1.5-7B across different pruning ratios. ``Average'' denotes the mean relative performance retention compared to the vanilla model. ``Last Token'' reports the number of visual tokens remaining in the final layer, highlighting the aggressive reduction of our trend-aware pruning approach compared to static methods. ``FLOPs'' measures the computational cost.}
        \resizebox{\linewidth}{!}{
                \begin{tabular}{ccccccccc|c|c|c}
                \hline
                \multicolumn{1}{c|}{Method} & \multicolumn{1}{c|}{Present at} & MME & GQA & POPE & SQA & RWQA & MMB & VizWiz & Average (\%) & Last Token & FLOPs \\ \hline
                \multicolumn{12}{c}{\cellcolor[HTML]{EFEFEF}\textit{Upper Bound, 576 Tokens}} \\ \hline
                \multicolumn{1}{c|}{LLaVA-v1.5-7B} & \multicolumn{1}{c|}{NeurIPS’24} & 1855.43 & 61.95 & 86.98 & 69.56 & 55.80 & 67.00 & 54.08 & 100\% & 576 & 100\% \\ \hline
                \multicolumn{12}{c}{\cellcolor[HTML]{EFEFEF} \textit{Retain 288 Tokens (↓ 50\%)}} \\ \hline
                \multicolumn{1}{c|}{FastV} & \multicolumn{1}{c|}{ECCV'24} & 1869.87 & 60.32 & 85.09 & 68.77 & 53.59 & 63.74 & 54.11 & 98.01 (↓1.99)  & 259 & 60.54\% \\
                \multicolumn{1}{c|}{PDrop} & \multicolumn{1}{c|}{CVPR'25} & 1846.67 & 60.74 & 86.83 & 68.02 & 53.73 & 64.34 & 53.61 & 98.09 (↓1.91)  & 178 & 63.49\% \\
                \multicolumn{1}{c|}{FasterVLM} & \multicolumn{1}{c|}{ICCV'25} & 1793.56 & 60.61 & 87.19 & 68.57 & 53.07 & 64.09 & 53.60 & 97.60 (↓2.40)  & 288 & 58.05\% \\
                \multicolumn{1}{c|}{SparseVLM} & \multicolumn{1}{c|}{ICML'25} & 1851.54 & 61.24 & 87.01 & 68.62 & 53.73 & 64.17 & 53.49 & 98.33 (↓1.67)  & 172 & 60.09\% \\
                \multicolumn{1}{c|}{VisionZip} & \multicolumn{1}{c|}{CVPR'25} & 1794.37 & 60.31 & 87.07 & 68.57 & 52.81 & 64.17 & 53.56 & 97.46 (↓2.54)  & 288 & 63.27\% \\
                \rowcolor{red!5} 
                \multicolumn{1}{c|}{Ours} & \multicolumn{1}{c|}{-} & 1867.10 & 61.53 & 86.64 & 69.21 & 54.64 & 64.18 & 53.80 & \textbf{98.89 (↓1.11)}  & $\sim$146 & \textbf{55.10\%} \\ \hline
                \multicolumn{12}{c}{\cellcolor[HTML]{EFEFEF}\textit{Retain 192 Tokens (↓ 66.7\%) }} \\ \hline
                \multicolumn{1}{c|}{FastV} & \multicolumn{1}{c|}{ECCV'24} & 1789.72 & 57.84 & 81.56 & 69.31 & 52.94 & 62.97 & 54.91 & 96.23 (↓3.77)  & 152 & 51.93\% \\
                \multicolumn{1}{c|}{PDrop} & \multicolumn{1}{c|}{CVPR'25} & 1802.14 & 59.27 & 86.10 & 68.86 & 52.55 & 64.00 & 53.35 & 97.02 (↓2.98)  & 49 & 45.12\% \\
                \multicolumn{1}{c|}{FasterVLM} & \multicolumn{1}{c|}{ICCV'25} & 1779.92 & 59.26 & 86.41 & 68.47 & 52.68 & 63.83 & 54.08 & 97.01 (↓2.99)  & 192 & 44.67\% \\
                \multicolumn{1}{c|}{SparseVLM} & \multicolumn{1}{c|}{ICML'25} & 1794.44 & 59.45 & 86.68 & 68.86 & 53.07 & 63.83 & 54.48 & 97.49 (↓2.51)  & 110 & 44.22\% \\
                \multicolumn{1}{c|}{VisionZip} & \multicolumn{1}{c|}{CVPR'25} & 1779.03 & 59.26 & 86.38 & 68.67 & 52.03 & 62.20 & 53.95 & 96.49 (↓3.51)  & 192 & 49.66\% \\
                \rowcolor{red!5} 
                \multicolumn{1}{c|}{Ours} & \multicolumn{1}{c|}{-} & 1849.73 & 59.36 & 84.92 & 68.86 & 54.64 & 63.23 & 54.16 & \textbf{97.80 (↓2.20)}  & $\sim$73 & \textbf{42.63\%} \\ \hline
                \multicolumn{12}{c}{\cellcolor[HTML]{EFEFEF} \textit{Retain 128 Tokens (↓ 77.8\%)} } \\ \hline
                \multicolumn{1}{c|}{FastV} & \multicolumn{1}{c|}{ECCV'24} & 1779.15 & 56.83 & 80.24 & 68.77 & 52.94 & 62.46 & 55.08 & 95.53  (↓4.47)  & 80 & 32.43\% \\
                \multicolumn{1}{c|}{PDrop} & \multicolumn{1}{c|}{CVPR'25} & 1752.40 & 59.06 & 81.01 & 68.72 & 51.90 & 63.32 & 53.07 & 95.34 (↓4.66)  & 45 & 36.28\% \\
                \multicolumn{1}{c|}{FasterVLM} & \multicolumn{1}{c|}{ICCV'25} & 1770.10 & 57.85 & 84.46 & 68.47 & 52.03 & 62.46 & 54.30 & 95.88 (↓4.12)  & 128 & 35.15\% \\
                \multicolumn{1}{c|}{SparseVLM} & \multicolumn{1}{c|}{ICML'25} & 1752.49 & 58.43 & 86.29 & 68.52 & 52.03 & 63.66 & 53.68 & \textbf{96.29 (↓3.71) } & 36 & 37.41\% \\
                \multicolumn{1}{c|}{VisionZip} & \multicolumn{1}{c|}{CVPR'25} & 1767.25 & 57.67 & 84.61 & 68.82 & 51.90 & 62.20 & 54.10 & 95.78 (↓4.22)  & 128 & 32.43\% \\
                \rowcolor{red!5} 
                \multicolumn{1}{c|}{Ours} & \multicolumn{1}{c|}{-} & 1812.35 & 57.52 & 82.34 & 69.96 & 52.42 & 62.29 & 53.83 & 96.03 (↓3.97)  & $\sim$23 & \textbf{32.20\%} \\ \hline
                \end{tabular}
        }

        \label{tab:mian_result}
\end{table*}

\subsection{Flow Activation and Dynamic Retention}
\label{subsec-activation}

Flow Activation unifies coarse static pruning with fine-grained trend-aware activation, enabling the model to decide which tokens to keep not only based on where they are, but also on where they are going.
The Flow Activation module determines the final set of retained tokens through a complementary integration strategy that unifies static attention-based pruning and dynamic trend-aware activation. Concretely, we first perform a Top-$k$ pruning based on layer-wise attention to remove evidently redundant tokens, serving as a coarse filtering stage for computational efficiency. On top of this static selection, we introduce a trend-aware mechanism that explicitly models the cross-layer evolution of token importance.

Unlike conventional pruning approaches that make irreversible decisions from a single layer, Flow Activation accounts for the potential future contribution of each token. Tokens that are weakly activated in the current layer $l$ but exhibit emerging or complementary trends across layers are selectively reactivated. The final retention set $\mathcal{S}^l_{\text{final}}$ is the union:
\begin{equation}
    \mathcal{S}^l_{\text{final}} = \mathcal{S}^l_{\text{rank}} \cup \mathcal{S}^l_{\text{up}} \cup \mathcal{S}^l_{\text{down}} \cup \mathcal{S}^l_{\text{fluct}},
\end{equation}
where  $\mathcal{S}^l_{\text{rank}}$ denotes the indices of the Top-$k$ pruning retained tokens. $ \mathcal{S}^l_{\text{up}} $, $\mathcal{S}^l_{\text{down}} $ and $\mathcal{S}^l_{\text{fluct}}$ represent the indices of the adaptive flow identification retained tokens.

This design enables the recovery of semantically critical but prematurely pruned tokens, facilitating the propagation of more robust and informative vision cues. By jointly leveraging the current layer attention and historical token trends, Flow Activation not only reduces redundant token processing, but also preserves the relevance of vision representations. 
As a result, it provides a coarse-to-fine and adaptive mechanism for context-aware token selection, preserving representational completeness.

\subsection{Computational Complexity Analysis}
\label{subsec-TheoreticalAnalysis}

We analyze the computational complexity in terms of floating-point operations (FLOPs) and demonstrate that our trend-aware pruning reduces the overall inference cost.
For a standard transformer decoder layer $l$ with hidden dimension $d$ and input sequence length $N = N_l^{v} + N_l^{t}$, the primary computation arises from the multi-head self-attention (MHSA) and feed-forward network (FFN) modules:
\begin{equation}
\mathcal{C}_{\text{base}}^{(l)} = 4Nd^2 + 2N^2d + 2Ndm,
\end{equation}
where $m$ denotes the intermediate size of the FFN.

In our framework, token pruning is performed before the execution of each decoder layer, starting from the second layer. Let $\hat{N}_l ={|\mathcal{S}^l_{\text{final}}|} + N_l^{t}$ denote the number of retained tokens after trend-aware pruning, the per-layer computational complexity is:
\begin{equation}
\mathcal{C}_{\text{ours}}^{(l)} = 4\hat{N}_ld^2 + 2\hat{N}{_l}^2d + 2\hat{N}_ldm.
\end{equation}

By reducing the token sequence before transformer decoder layers, our method reduces computational complexity in linear projection, attention mechanism, and FFN costs.
The additional overhead time complexity is introduced by trend-aware pruning, including element-wise operation, Top-$k$ selection, and index reordering. 
This overhead scales as $\mathcal{O}(W\hat{N}^v + \hat{N}^v \log \hat{N}^v)$, where $\hat{N}^v = |\mathcal{S}_{\text{final}}^l|$ represents the number of retained visual tokens, and W denotes the window size. In practice, this overhead is negligible compared to $\mathcal{O}({\hat{N}^v}d^2)$.

\section{Experiments}

\subsection{Experimental Settings}

We evaluate our method across diverse MLLM architectures, including the LLaVA series~\cite{llava,llavenext,llavaOneVision} and Qwen2.5-VL~\cite{qwenvl25}. Experiments are conducted on a wide range of standard benchmarks, including MME~\cite{mme}, GQA~\cite{gqa}, POPE~\cite{pope}, SQA~\cite{sqa}, MMBench~\cite{mmb}, VizWiz~\cite{VizWiz}, OCRBench~\cite{OCRBench}, InfoVQA~\cite{InfographicVQA} and AI2D~\cite{AI2D}. All methods are evaluated following their default settings. To ensure a fair comparison, all experiments are executed under an identical experimental setup and computational environment. All experiments are conducted on NVIDIA A100 GPUs. The implementation is carried out in Python 3.10, utilizing PyTorch 2.1.2, and transformers 4.37.2. To ensure fair comparison, we strictly follow the official inference settings for all baseline models without modification, including LLaVA-v1.5, LLaVA-NeXT, LLaVA-OneVision, and Qwen2.5-VL.

\subsection{Main Results}

We evaluate our trend-aware dynamic pruning method on LLaVA-v1.5-7B across different sparsity level configurations and compare against representative training-free token pruning methods, as present in \cref{tab:mian_result}.
At a 50\% pruning ratio, our method achieves the highest average performance retention 98.89\% while also delivering the lowest FLOPs 55.10\%, outperforming FastV (98.01\%, 60.54\% FLOPs), PDrop (98.09\%, 63.49\% FLOPs), and SparseVLM (98.33\%, 60.09\% FLOPs). Meanwhile, our approach reduces the final-layer visual tokens to $\sim$146, substantially fewer than most methods (e.g., FastV: 259, SparseVLM: 172). When the pruning ratio increases to 66.7\%, Trend-aware Pruning maintains 97.80\% performance retention, surpassing PDrop (97.02\%), and VisionZip (96.49\%), while achieving the lowest computational cost (42.63\% FLOPs) and compressing the tokens to $\sim$73, compared to SparseVLM’s 110 and FasterVLM’s 192 tokens. 
Under the most aggressive setting (77.8\%), although SparseVLM attains slightly higher average accuracy (96.29\% vs. 96.03\%), our method drastically reduces the final-layer visual tokens to only $\sim$23, the smallest among all approaches, while still achieving the lowest FLOPs (32.20\%). 
These results demonstrate that modeling token importance as a dynamic, trend-aware process enables more precise and stable pruning decisions than static methods, effectively removing redundant visual tokens while preserving task-critical information.

\begin{table}[!t]
\centering
\caption{Performance comparison of three MLLM architectures under varying token retention ratios.}
\resizebox{\linewidth}{!}{
\begin{tabular}{c|c|cccccc|c}
\toprule
Method & Retention & MME & POPE & MMB & SQA & AI2D & GQA & Average \\ \midrule
LLaVA-v1.5-13B & 100\% & 1817.95 & 87.08 & 68.81 & 72.78 & 59.26 & 63.29 & 100\% \\ \hline
FastV &  & 1804.09 & 85.87 & 68.47 & 73.33 & 58.55 & 62.16 & 99.19\% \\
\cellcolor[HTML]{EFEFEF}Ours & \multirow{-2}{*}{45\%} & \cellcolor[HTML]{EFEFEF}1830.69 & \cellcolor[HTML]{EFEFEF}87.12 & \cellcolor[HTML]{EFEFEF}68.38 & \cellcolor[HTML]{EFEFEF}73.62 & \cellcolor[HTML]{EFEFEF}58.87 & \cellcolor[HTML]{EFEFEF}62.94 & \cellcolor[HTML]{EFEFEF}\textbf{100.01\%} \\ \hline
FastV &  & 1759.75 & 81.67 & 66.84 & 74.17 & 57.25 & 58.93 & 96.56\% \\
\cellcolor[HTML]{EFEFEF}Ours & \multirow{-2}{*}{20\%} & \cellcolor[HTML]{EFEFEF}1823.39 & \cellcolor[HTML]{EFEFEF}86.61 & \cellcolor[HTML]{EFEFEF}67.53 & \cellcolor[HTML]{EFEFEF}73.72 & \cellcolor[HTML]{EFEFEF}58.19 & \cellcolor[HTML]{EFEFEF}61.58 & \cellcolor[HTML]{EFEFEF}\textbf{99.11\%} \\ \hline
LLaVA-Next-7B & 100\% & 1825.54 & 87.89 & 68.13 & 72.98 & 67.36 & 64.82 & 100\% \\ \hline
FastV &  & 1813.98 & 87.51 & 67.18 & 72.93 & 66.61 & 63.79 & 99.13\% \\
\cellcolor[HTML]{EFEFEF}Ours & \multirow{-2}{*}{45\%} & \cellcolor[HTML]{EFEFEF}1831.49 & \cellcolor[HTML]{EFEFEF}87.86 & \cellcolor[HTML]{EFEFEF}67.96 & \cellcolor[HTML]{EFEFEF}72.83 & \cellcolor[HTML]{EFEFEF}66.26 & \cellcolor[HTML]{EFEFEF}64.31 & \cellcolor[HTML]{EFEFEF}\textbf{99.57\%} \\ \hline
FastV &  & 1780.50 & 86.84 & 65.89 & 73.08 & 66.42 & 62.95 & 98.15\% \\
\cellcolor[HTML]{EFEFEF}Ours & \multirow{-2}{*}{33\%} & \cellcolor[HTML]{EFEFEF}1844.71 & \cellcolor[HTML]{EFEFEF}87.76 & \cellcolor[HTML]{EFEFEF}67.87 & \cellcolor[HTML]{EFEFEF}72.98 & \cellcolor[HTML]{EFEFEF}66.13 & \cellcolor[HTML]{EFEFEF}64.03 & \cellcolor[HTML]{EFEFEF}\textbf{99.58\%} \\ \hline
LLaVA-OV-0.5B & 100\% & 1478.41 & 88.29 & 52.06 & 67.18 & 57.09 & 57.95 & 100\% \\ \hline
FastV &  & 1531.70 & 83.58 & 50.43 & 66.19 & 55.21 & 52.65 & 96.87\% \\
\cellcolor[HTML]{EFEFEF}Ours & \multirow{-2}{*}{50\%} & \cellcolor[HTML]{EFEFEF}1520.15 & \cellcolor[HTML]{EFEFEF}85.63 & \cellcolor[HTML]{EFEFEF}50.95 & \cellcolor[HTML]{EFEFEF}66.14 & \cellcolor[HTML]{EFEFEF}55.63 & \cellcolor[HTML]{EFEFEF}54.09 & \cellcolor[HTML]{EFEFEF}\textbf{97.82\%} \\ \hline
FastV &  & 1423.23 & 74.94 & 47.16 & 64.75 & 53.11 & 47.20 & 90.43\% \\
\cellcolor[HTML]{EFEFEF}Ours & \multirow{-2}{*}{20\%} & \cellcolor[HTML]{EFEFEF}1470.25 & \cellcolor[HTML]{EFEFEF}77.59 & \cellcolor[HTML]{EFEFEF}47.25 & \cellcolor[HTML]{EFEFEF}65.54 & \cellcolor[HTML]{EFEFEF}52.40 & \cellcolor[HTML]{EFEFEF}47.53 & \cellcolor[HTML]{EFEFEF}\textbf{91.57\%} \\ \hline
\end{tabular}
}

\label{tab:gen-results}

\end{table}

\subsection{Performance on More Advanced MLLMs}
\noindent \textbf{Evaluation on LLaVA Series.}
As shown in \cref{tab:gen-results}, our proposed Trend-aware Pruning demonstrates applicability across the LLaVA series. For instance, with 45\% visual tokens retained on LLaVA-v1.5-13B, our method maintains a high average of 100.01\%, marginally exceeding the full-token baseline which suggests the substantial reduction of redundant visual noise. At 20\% retention, our method surpasses FastV by 2.55\% on LLaVA-v1.5-13B and 1.14\% on LLaVA-OV-0.5B. Similarly, on LLaVA-Next-7B with 33\% retention, our method outperforms FastV by 1.4\%, demonstrating consistent superiority across varying model capacities.

\begin{table}[!t]
\centering
\setlength{\tabcolsep}{1.5pt}
\renewcommand{\arraystretch}{1.1}
    \caption{Performance comparison of static and our method Trend-aware Pruning on Qwen2.5-VL.}
            \label{tab:qwen}

        \resizebox{\linewidth}{!}{
\begin{tabular}{c|l|ccccccc}
\hline
Model & \multicolumn{1}{c|}{Method} & MME  & POPE & MMB & SQA & AI2D & GQA\\ \hline
 & \begin{tabular}[c]{@{}l@{}}+ Top-$k$ Pruning\end{tabular}  & 1612.47  & 71.86 & 51.55  & 74.86 & 67.75 & 44.23 \\ \cline{2-2}
 \multirow{-3}{*}{Qwen2.5-VL} & \cellcolor[HTML]{EFEFEF}\begin{tabular}[c]{@{}l@{}}+ Trend-aware  \\ Pruning\end{tabular} & \cellcolor[HTML]{EFEFEF}1866.70  & \cellcolor[HTML]{EFEFEF}81.89 & \cellcolor[HTML]{EFEFEF}66.24  & \cellcolor[HTML]{EFEFEF}77.39 & \cellcolor[HTML]{EFEFEF}68.94 & \cellcolor[HTML]{EFEFEF}49.15\\ \hline
\end{tabular}         }

\end{table}

\noindent \textbf{Generalization to Qwen2.5-VL.}
To further verify the universality of our Trend-aware Pruning, we extend our evaluation from the LLaVA series to the Qwen2.5-VL. Our method demonstrates a substantial performance advantage over the static Top-$k$ Pruning baseline, with the results  presented in \cref{tab:qwen}. To ensure a fair comparison, we employ a consistent Top-$k$ pruning baseline across layers, distinguishing our method solely by the incorporation of trend-aware retained tokens to capture dynamic visual cues. Notably, on the MMB and POPE tasks, our approach achieves scores of 66.24 and 81.89, significantly outperforming the Top-$k$ Pruning by margins of 14.69 and 10.03 points, respectively. This evidence suggests that simple magnitude-based pruning struggles to retain semantic integrity in advanced architectures, whereas our trend-aware strategy effectively preserves critical visual cues regardless of the underlying model structure.

\begin{table}[!t] 
    \centering
    \small
        \caption{Ablation study on flow identification strategies for token retention. We evaluate the impact of progressively incorporating Upward, Fluctuating and Downward trending tokens.}
        
        \resizebox{\linewidth}{!}{
            \begin{tabular}{l|cccccc}
            \hline
            Retained & MMB & GQA  & VizWiz & OCRBench & InfoVQA \\ \hline
            Top-$k$ &  63.23 & 60.13 & 53.44 & 18.70 & 19.57 \\ \hline
            \begin{tabular}[c]{@{}l@{}}Top-$k$\\ +Upward\end{tabular}& 63.57 & 60.97 & 53.50 & 25.30 & 19.85\\ \hline
            \begin{tabular}[c]{@{}l@{}}Top-$k$ \\ + Upward  \\ + Fluctuating\end{tabular} & 63.57 & 60.97 & 53.55 & 25.70 & 19.91 \\ \hline
            \cellcolor[HTML]{EFEFEF}\begin{tabular}[c]{@{}l@{}} Top-$k$ \\ + Upward  \\ + Fluctuating \\ + Downward\end{tabular} & \cellcolor[HTML]{EFEFEF}64.18 & \cellcolor[HTML]{EFEFEF}61.53  & \cellcolor[HTML]{EFEFEF}53.80 & \cellcolor[HTML]{EFEFEF}31.00   & \cellcolor[HTML]{EFEFEF}20.09\\ \hline
            \end{tabular}
        }
        \label{tab:flowidentification}
    
\end{table}

\subsection{Ablation Study}

\noindent \textbf{Effectiveness of Flow Identification.}
To explicitly validate whether our dynamic retention effectively preserves delicate visual cues, we stress-test our method on detail-heavy and OCR-centric suites, specifically OCRBench and InfoVQA (\cref{tab:flowidentification}). Small, fine-grained objects often exhibit weak attention in shallow layers and are highly vulnerable to premature discarding.
As shown in \cref{tab:flowidentification}, relying solely on static Top-$k$ pruning results in severe semantic degradation on these tasks (e.g., scoring a mere 18.70 on OCRBench). Based on the trend-aware modeling, the initial incorporation of Upward-trending tokens results in a significant performance gain, notably increasing the OCRBench score by 6.60 points. Subsequently, the addition of Fluctuating tokens offers further incremental refinement to the model's accuracy. Crucially, the final integration of Downward-trending tokens maximizes performance, achieving scores of 31.00 on OCRBench and 20.09 on InfoVQA.
This compelling evidence demonstrates that our trend-aware modeling successfully preserves fine-grained visual features that static pruning fails to capture.

\begin{figure}[!t]
\begin{center}
\centerline{\includegraphics[width=0.95\columnwidth]{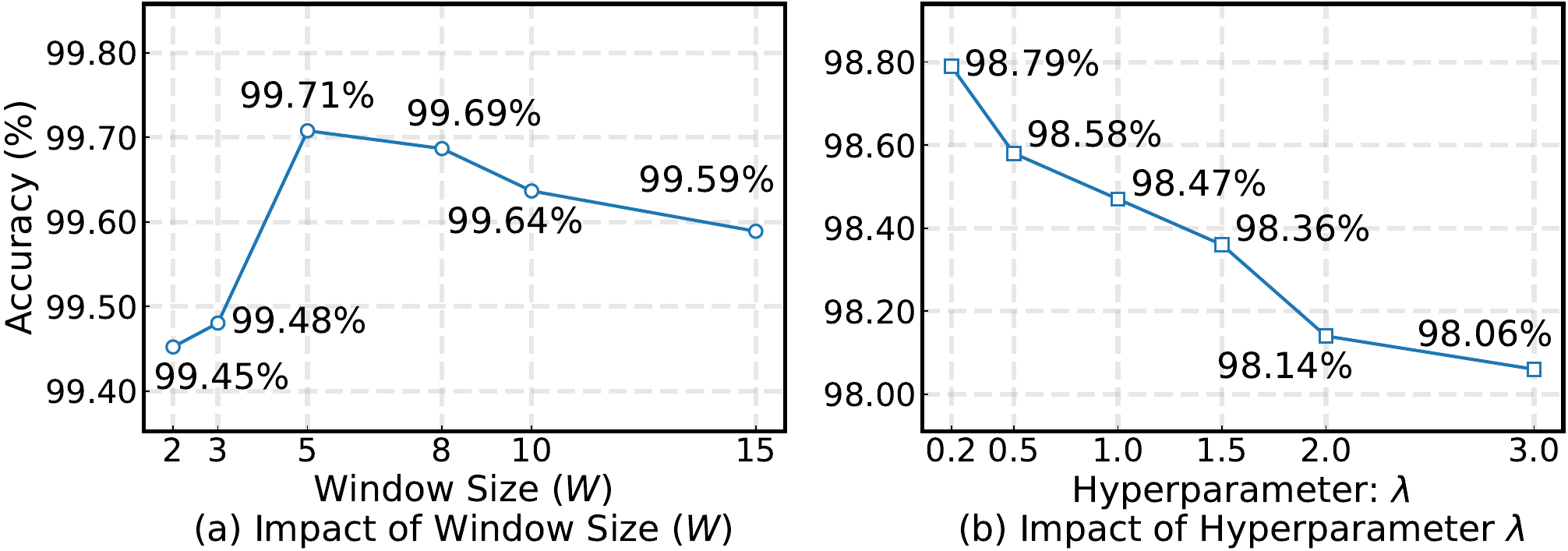}}
\caption{Ablation study on hyperparameter sensitivity. 
}
\label{window_size_ijcai}
\end{center}
\vskip -0.3in
\end{figure}

\noindent \textbf{Impact of Window Size and $\lambda$.} We evaluate the sensitivity of our method to the window size $W$ and threshold $\lambda$, reporting the average accuracy across five representative benchmarks (MME, GQA, POPE, SQA, and VizWiz).
As shown in \cref{window_size_ijcai} (a), the performance improves as the window size increases, peaking at $W=5$ with an accuracy of 99.71\%. We observe that smaller window sizes fail to capture sufficient historical context to identify meaningful trends. Conversely, larger windows (e.g., $W \ge 10$) incorporate irrelevant history, which leads to a performance drop. Regarding $\lambda$, \cref{window_size_ijcai} (b) demonstrates that the model maintains stable performance within the range of $0.2 \le \lambda \le 1.0$. Based on these empirical results, we adopt $W=5$ and $\lambda=0.5$ as the default settings, which achieve a favorable balance between effective trend capture, robustness, and controllable pruning behavior.

\section{Conclusion}

We presented \textit{Trend-aware Pruning}, a training-free framework that models the cross-layer evolution of visual tokens in MLLMs and formulates pruning as a dynamic, reversible process. By capturing token-level flow tendencies, our method adaptively prunes and recovers tokens, preserving semantically critical visual cues while substantially reducing computation. Extensive experiments demonstrate competitive or superior performance across diverse architectures and benchmarks. Beyond efficiency gains, Trend-aware Pruning offers a new perspective on visual token pruning, highlighting the importance of modeling token dynamics across layers and opening promising directions for future research on reversible and representation-consistent pruning strategies.

\noindent \textbf{Limitations and Future Work.} While training-free and broadly applicable, Trend-aware Pruning currently relies on lightweight cross-layer statistics with a fixed observation window, which may not fully capture long-range token dependencies or higher-level visual structures. 
Future work could explore more expressive trend modeling, extend pruning from token-level to region- or concept-level representations, and investigate joint optimization with lightweight adaptation. The Token Collector could build on adaptive window mechanisms conditioned on the input sequence length.
Moreover, extending this approach to video understanding and long-horizon multimodal reasoning remains an exciting direction.

\begin{acks}
This work was supported by the New Generation Artificial Intelligence-National Science and Technology Major Project (No. 2025ZD0122701), the National Natural Science Foundation of China (No. U22B2051, No. U25B2066, No. 62302411), the Xiamen Municipal Science and Technology Bureau, China (3502ZC-QXT2024009).
\end{acks}

\bibliographystyle{ACM-Reference-Format}
\bibliography{sample-base}










\end{document}